\documentclass[lettersize,journal]{IEEEtran}
\usepackage{array}
\usepackage[caption=false,font=normalsize,labelfont=sf,textfont=sf]{subfig}
\usepackage{stfloats}
\usepackage{url}
\usepackage{verbatim}
\usepackage{float}

\usepackage{atbegshi,picture}
\usepackage{lipsum}

\usepackage{cite}
\usepackage{amsmath,amssymb,amsfonts}
\usepackage{gensymb}
\usepackage{algorithmic}
\usepackage{subcaption}
\usepackage{graphicx}
\usepackage{textcomp}
\usepackage{xcolor}
\usepackage{cleveref}
\usepackage[inline]{enumitem}
\usepackage{booktabs}

\usepackage[detect-none]{siunitx}
\newcommand{\mAh}{mAh}
\newcommand{\mps}{\meter / \second}
\usepackage{glossaries}
\newacronym{tof}{ToF}{time-of-flight}
\newacronym{ekf}{EKF}{extended Kalman filter}
\newacronym{fov}{FoV}{field-of-view}
\newacronym{soc}{SoC}{system-on-chip}
\newacronym[plural=MCUs,firstplural=microcontroller units (MCUs)]{mcu}{MCU}{microcontroller unit}
\newacronym[plural=CNNs,firstplural=convolutional neural networks (CNNs)]{cnn}{CNN}{convolutional neural network}
\newacronym{slam}{SLAM}{simultaneous localization and mapping}
\newacronym{lidar}{LiDAR}{light detection and ranging}
\newacronym{radar}{RADAR}{Radio Detection and Ranging}
\newacronym{iqd}{IQD}{interquartile deviation}
\newacronym{bce}{BCE}{binary cross-entropy}
\newacronym{auroc}{AUROC}{area under receiver operating characteristic curve}
\newacronym{rmse}{RMSE}{root mean squared error}
\newacronym{uart}{UART}{universal asynchronous receiver/transmitter}
\newacronym{imav}{IMAV}{International Micro Air Vehicles}
\newacronym{soa}{SoA}{state of the art}
\newacronym{uav}{UAV}{unmanned aerial vehicle}
\newacronym{vml}{VML}{visual model‐predictive localization}
\newacronym{dnn-rl}{DNN-RL}{deep neural network reinforcement learning}
\newacronym{ssd}{SSD}{single-shot multibox detector}
\newacronym{vio}{VIO}{visual-inertial odometry}
\newacronym{qat}{QAT}{quantization aware training}
\newacronym{oa}{OA}{obstacle avoidance}
\newacronym{cf}{CF}{Crazyflie 2.1}
\newacronym{us}{US}{ultrasound}
\newacronym{aln}{AlN}{Aluminum Nitride} 
\newacronym{pmut}{PMUT}{piezoelectric micromachined ultrasonic transducer} 
\newacronym{mems}{MEMS}{micro-electromechanical systems}
\newacronym{pulp}{PULP}{Parallel Ultra-Low-Power}
\newacronym{odr}{ODR}{output data rate}
\newacronym{iq}{IQ}{in-phase and quadrature}
\newacronym{snr}{SNR}{signal-to-noise ratio}
\newacronym{imu}{IMU}{inertial measurement unit}
\newacronym{mse}{MSE}{mean square error}
\begin{document}

\title{BatDeck - Ultra Low-power Ultrasonic Ego-velocity Estimation and Obstacle Avoidance on Nano-drones}

\author{\IEEEauthorblockN{Hanna M\"uller$^{1}$, Victor Kartsch$^{1}$, Michele Magno$^{1}$, Luca Benini$^{1,2}$}
        % <-this % stops a space
\thanks{\IEEEauthorblockA{$^{1}$Integrated Systems Laboratory / Center for Project-Based Learning  - ETH Z\"urich, Switzerland}

\IEEEauthorblockA{$^{2}$DEI - University of Bologna, Italy}

Email: \{hanmuell, vkartsch, mmagno, lbenini\}@ethz.ch
}
}

% \AtBeginShipout{\AtBeginShipoutUpperLeft{%
%   \put(\dimexpr\paperwidth-1cm\relax,-1.5cm){\makebox[0pt][r]{\framebox{1571011846}}}%

\maketitle

\begin{abstract}
Nano-drones, with their small, lightweight design, are ideal for confined-space rescue missions and inherently safe for human interaction. However, their limited payload restricts the critical sensing needed for ego-velocity estimation and obstacle detection to single-bean laser-based time-of-flight (ToF) and low-resolution optical sensors. Although those sensors have demonstrated good performance, they fail in some complex real-world scenarios, especially when facing transparent or reflective surfaces (ToFs) or when lacking visual features (optical-flow sensors). Taking inspiration from bats, this paper proposes a novel two-way ranging-based method for ego-velocity estimation and obstacle avoidance based on down-and-forward facing ultra-low-power ultrasonic sensors, which improve the performance when the drone faces reflective materials or navigates in complete darkness. Our results demonstrate that our new sensing system achieves a mean square error of 0.019 m/s on ego-velocity estimation and allows exploration for a flight time of 8 minutes while covering 136 m on average in a challenging environment with transparent and reflective obstacles. We also compare ultrasonic and laser-based ToF sensing techniques for obstacle avoidance, as well as optical flow and ultrasonic-based techniques for ego-velocity estimation,  denoting how these systems and methods can be complemented to enhance the robustness of nano-drone operations.

\end{abstract}

\begin{IEEEkeywords}
Drone, state estimation, nano-UAV, ultrasonic.
\end{IEEEkeywords}

\section{Introduction}
\IEEEPARstart{T}{he} projected growth in the UAV market through 2030 reflects a rising interest among businesses and individuals in the diverse solutions that UAV technology can enable in different domains such as agriculture, public safety, urban planning, and entertainment\cite{fortune_business_insights_2024}. Specifically, nano-drones (typically $\sim$\qty{10}{\centi\meter} diameter~\cite{HASSANALIAN201799})  allow exploring constrained environments like greenhouses and buildings, while being safe for people and property. 

Drones require precise state information—including position, velocity, and orientation—to execute missions reliably. Achieving precision involves algorithms that integrate data from multiple sensors. Algorithms like \gls{slam} \cite{macario2022comprehensive} and \gls{vio} \cite{he2020review} combine visual data (from cameras) and inertial data to continuously track a drone's position and orientation. The \gls{ekf} offers a computationally efficient approach to resource-limited platforms, such as nano-drones, by fusing data from various sensors, including \glspl{imu} and low-resolution optical flow sensors. Even though there have been advances in optical flow robustness in recent years~\cite{tim2022egomotion,tim2021monocular}, estimates can be severely affected when optical conditions for optical flow are suboptimal, such as in low-light or feature-less settings.

Ultrasound, which measures the \gls{tof} in echos of high-frequency sound, has the potential for providing robust state estimation, even without optical features~\cite{oho2022ultrasonic}. A wide range of ultrasonic sensors applicable for use in robotics are currently available on the market and have been examined in the literature\cite{zhmud2018application, rshen2019new, tim2011indoor, oho2022ultrasonic}. However, their size, weight, and energy requirements preclude their use in nano-drone applications.

\begin{figure}
\centering
\includegraphics[width=0.95\columnwidth]{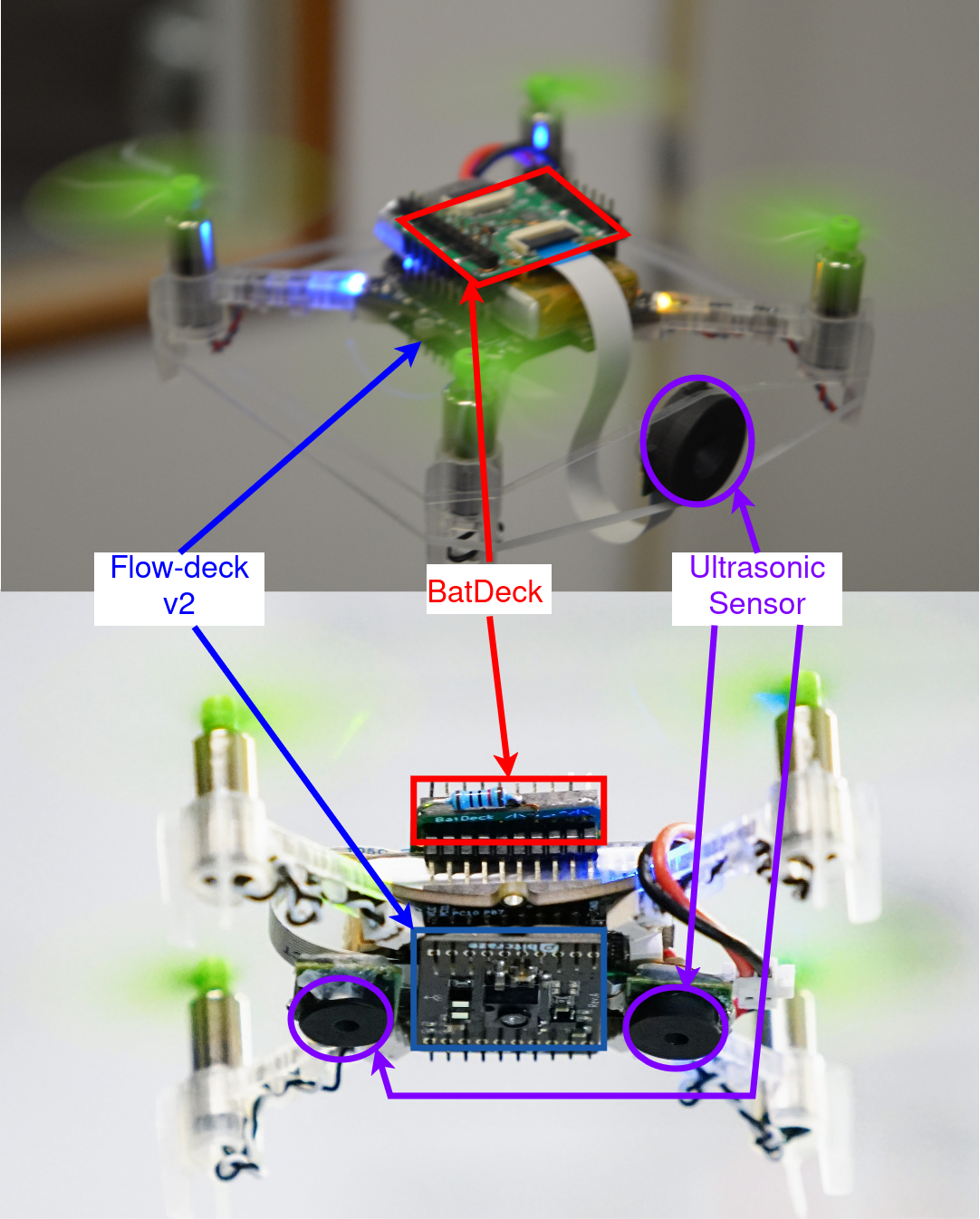}
\caption{The Crazyflie 2.1 with the \textit{BatDeck} and the Flow-deck v2, equipped for OA (top) and ego-velocity estimation (bottom).}
\label{fig:hw_overview}
\end{figure}

The newly introduced ICU-x0201 sensors from TDK (as shown in our setup in \Cref{fig:hw_overview}) overcome these limitations with their ultra-low power consumption (below \qty{1}{\milli\watt}) and compact form factor (\qty{5.17}{\times}\qty{2.68}{\times}\qty{0.9}{\milli\meter}), making them an ideal choice for integration into small-scale robotic platforms~\cite{przybyla2023mass}. However, for using a Doppler-shift-based velocity estimation as used in \cite{oho2022ultrasonic}, a highly directed ultrasonic wave is needed, leading to bulky and heavy acoustic horns, which are infeasible on nano-drones. Hence there is a need to develop a novel ego-velocity method with ultrasonic sensors for nano-drones.

% obstacle avoidance
\Gls{oa} is also critical to ensure the drone platform's safe and efficient operation. Often, \gls{oa} is performed with high-resolution sensors as \gls{lidar}, \gls{radar} or cameras~\cite{tim2022multigoal, 10423569}. However, \gls{lidar} and \gls{radar} are bulky and power-hungry, and are thus unsuitable for nano-drones applications. Laser-based \gls{tof} sensors have also been featured in systems coupled with naturally-inspired \gls{oa} algorithms, providing a low-power and robust performance \cite{ mueller2023robust, lamberti2022tiny}. Still, camera and laser-based solutions are heavily impaired when dealing with reflective surfaces or transparent barriers such as glass walls. 

A viable solution to address these robustness challenges is to perform sensor fusion of multiple sensors, but nano-drones that are typically below \qty{50}{\gram} can only resort to a small and lightweight set of sensors given payload and power constraints~\cite{HASSANALIAN201799}. Inspired by the echolocation techniques used by bats and dolphins, ultrasonic sensors — which measure \gls{tof} using echoes of high-frequency sound — offer a promising alternative, excelling with sound-reflective objects regardless of color, transparency, or texture~\cite{laurijssen2019flexible,forouher2016sensor}.

This work explores equipping nano-drones with bat-like navigation capabilities by leveraging a novel, compact, and low-power ultrasonic sensor. It extends previous work\cite{muller2024batdeck} by introducing a novel method for ego-velocity estimation. Specifically, our contributions are:

\begin{enumerate}
    \item Design and implementation of an ultrasonic hardware extension system (PCB) for nano-drones (\textit{BatDeck}), integrating up to four compact and energy-efficient ICU-x0201 sensors that enable emitting and receiving ultrasonic signals.
     \item Implementation and evaluation of an \gls{oa} algorithm tailored for resource-constrained lightweight hardware (ARM Cortex-M4) onboard a nano-drone in terms of power and latency.
     \item In-field evaluation and comparison of \textit{BatDeck} with laser-based \gls{tof} sensors for \gls{oa} applications, based on mission success rate, power consumption, and adaptability to various environments, providing insights into the advantages and potential limitations of both approaches in a real-world scenario.
     \item Development and implementation, including the mathematical background, of a novel ego-velocity estimation algorithm relying on two ground-facing ultrasonic sensors. 
     \item Proof-of-concept in-field evaluation and comparison of the ultrasonic ego-velocity estimation to an optical-flow-based sensor for velocity estimation in different environments, highlighting how these sensors can be complemented to provide a more robust drone state estimation.
\end{enumerate}

\section{Related Work}

Ego-velocity estimation involves determining an object's speed and movement direction relative to its environment, essential for navigation, obstacle avoidance, and maintaining stability in dynamic settings. Numerous studies have tacked the estimation of accurate ego-motion parameters based on various sensing modalities. For instance, Shan et al. \cite{shan2020probabilistic} proposed a probabilistic framework utilizing lidar and \gls{imu} data to correct motion-induced distortions in lidar point clouds, thereby improving ego-velocity reliability. Camera-based systems, often when combined with \glspl{imu}, have proven effective for ego-motion and ego-velocity estimation as a part of \gls{slam}, \gls{vio}~\cite{macario2022comprehensive, he2020review} or \glspl{cnn}. In \cite{MUMUNI2022100416}, authors address ego-motion and depth estimation for UAV navigation in dynamic environments by introducing Cowan-GGR, a deep learning-based framework integrating optical flow, monocular depth, and IMU-based pose updates to enhance ego-motion estimation, particularly by using geometric relationships between depth and translation. Similarly, radar sensors have demonstrated reliability in velocity estimation; for instance, Park et al. \cite{park2021} introduced a mmWave Doppler radar system with two orthogonal radar modules to capture 3D motion, supporting accurate measurements in foggy or unstructured outdoor environments. However, these solutions are generally too bulky and power-intensive for application in nano-drones.

Cameras have also been used successfully for \gls{oa} tasks. Tiny, low-power cameras can easily be mounted and operated on small systems such as nano-drones. For instance, our group previously designed a vision-based \gls{cf}-compatible shield for \gls{oa}~\cite{palossi201964}. The device integrates a low-power monochrome QVGA camera (Himax HM01B0) and GAP8, an \gls{pulp} \gls{soc} for efficient computation. The navigational commands are based on PULP-DroNet, a convolutional neural network for \gls{oa} fine-tuned for this platform. Zhang et al.~\cite{zhang2024endtoend} use a lightweight CNN depth estimation network for \gls{oa} - the generated depth map could even be used for path planning. While camera-based approaches are compatible with nano-drones, they still lack robustness in poor lighting conditions or when navigating across an environment with objects that have not been included in the training sets.

Kahlenberg et al. have recently presented Stargate~\cite{10423569}, a multimodal sensor fusion system for nano-drones, enabling autonomous navigation and gate detection through sensor fusion between a low-power grayscale camera (Himax HM01B0) and a multi-pixel (8x8) laser-based \gls{tof} (VL53L5CX) sensor from STMicroelectronics. The system is trained entirely on synthetic data and tested on a \gls{cf} drone, demonstrating robust navigation with a low failure rate in various scenarios. While this approach works in many environments, its robustness can be improved by fusing an additional sensor perceiving glass or visually reflective materials correctly.

Cameras are not only commonly used for \gls{oa} alone but also for optical flow estimation, key for ego-velocity estimation, which helps to counter accumulated drift from \gls{imu} sensors\cite{barron1994performance}. However, also in this case, cameras are not robust in poor light conditions and on surfaces with a low amount of optical features.

Focusing on sound as an alternative sensing modality, Duembgen et al.~\cite{duembgen2023blind} developed a real-time, model-based auditory localization and mapping system designed explicitly for compact robots to exploit the capability that sounds offer to detect visually transparent and reflective obstacles. This system utilizes buzzers and simple microphones to achieve centimeter-level precision without prior calibration or training. Implementing this approach on nano-drones remains challenging due to the need for multiple components and strict limitations on weight and size. Recent advancements in MEMS design are nevertheless facilitating the development of low-power, lightweight, and compact components such as the ICU-x0201, a lightweight and low-power (\SI{<1}{\milli\watt}) ultrasonic \gls{tof} sensor, incorporating \gls{aln} \glspl{pmut} to offer enhanced sensitivity and reduced power consumption when compared to traditional piezoceramic transducers.

In \Cref{tab:sensing}, we summarize the benefits and disadvantages of cameras, laser-based and ultrasound-based sensors, showing that they can complement each other to develop a robust system for complex environments.

In this paper, we leverage the capabilities of the TDK ICU-x0201 sensors to enhance \textit{BatDeck}, an expansion board equipped with two downward-facing ultrasonic sensors and compatible with \gls{cf} drones, aimed at improving performance in \gls{oa} tasks. This work also introduces a methodology for ultrasound-based ego-velocity estimation, which operates effectively even in complete darkness by utilizing the phase shift between signals received by each sensor, complementing the sensing with optical sensors. We present a detailed characterization of the sensor under various flight conditions and across different materials, providing a quantitative assessment of its efficacy in complex environments compared to conventional laser-based \gls{tof} and visual optical flow sensors.

\begin{table*}[t]
\caption{Sensing technology comparison.}
\label{tab:sensing}
\centering
\footnotesize
\begin{tabular}{c|c|c|c|c|c}
\toprule
{\textbf{Technology}} & \textbf{Power consumption} & {\textbf{Darkness}} & \textbf{Reflective} & \textbf{Rough surface} & \textbf{Computational requirement}  \\ 
\midrule
Camera~\cite{palossi201964,zhang2024endtoend} & ++ & - - - & - - - & +++  & - -\\ \hline
Laser-based~\cite{mueller2023robust} & - & +++ & - - - & +++ & +++\\ \hline
Ultrasound~\cite{duembgen2023blind,oho2022ultrasonic}, \textbf{ours} & +++ & +++ & +++ & - & ++\\ \bottomrule
\end{tabular}
\end{table*}

\section{Background}
This section provides background on the used components.
\subsection{TDK ICU-x0201}
As ultrasonic sensors, we use the TDK ICU-x0201, a series of high-performance, miniature, ultra-low power, long-range ultrasonic \gls{tof} transceivers. \\
Two variants are used in this work, the ICU-30201, integrating a nominally $f_{op} =50kHz$ \gls{pmut} and a range of up to \SI{9}{\meter}, and the ICU-10201 which integrates a nominally 175 kHz \gls{pmut} and has a range of up to \SI{1.2}{\meter}. The \gls{fov} is configurable by mounting custom horns. Both sensors feature a \qty{40}{\mega\hertz} CPU for sampling and pre-processing.
They can record maximally 340 \gls{iq} data samples per measurement, converted to baseband. The \gls{odr} can be configured as $\gls{odr} = f_{op}/N$ where $N = 2,4,8$. The \gls{iq} allows computing phase and magnitude of the reflected ultrasonic waves. Choosing an \gls{odr} is a tradeoff between resolution and range - we use the best possible resolution, $N=2$ for the ego-velocity estimation, but choose $N=4$ for the \gls{oa} task to increase the range.

\subsection{BatDeck}
The \textit{BatDeck}, a custom-made extension deck for the Crazyflie, features four slots for attaching TDK ICU-x0201 sensors, giving us the flexibility to use the deck for both \gls{oa} and velocity estimation. 
% In this work, we only use one slot to attach a forward-facing ICU-30201 ultrasonic sensor with a \qty{55}{\degree} \gls{fov} horn, attached to the drone with two rubber bands (for dampening). 
The \textit{BatDeck} with one sensor weighs in total \qty{3}{\gram}, where the expansion deck PCB itself contributes 1.37 g, the sensor with horn \qty{1.25}{\gram}. For the dual sensor configuration, we add a second sensor with horn, totaling at \SI{\sim4.25}{\gram}

\subsection{Crazyflie Platform}
This work uses the \gls{cf} platform from Bitcraze, a 10-centimeter nano-drone featuring an STM32F405 \gls{mcu}, which oversees state estimation and actuation control. The platform is expanded with a Flow-deck v2, which enables optical flow (PMW3901) and height measurements to improve state estimation, based on an \gls{ekf}. In addition, we use 7$\times$\qty{16}{\milli\meter} 19000 KV motors from betafpv, 47-17 propeller from Bitcraze, and an \qty{350}{\mAh} LiPo battery from Tattu. This base configuration weighs \qty{34}{\gram}.

\subsection{Obstacle avoidance algorithm}
\begin{figure}
\centering
\includegraphics[width=0.5\columnwidth]{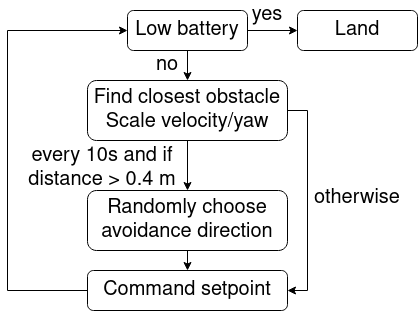}
\caption{Algorithmic flow of the \gls{oa} implementation~\cite{muller2024batdeck}.}
\label{fig:flow}
\end{figure}
The \gls{oa} algorithm is shown in \Cref{fig:flow}; it finds the distance to the closest object with the help of a dynamic threshold and then scales the forward velocity and the yaw angle accordingly. For random exploration, the yaw rate sign (avoiding obstacles to the left or right) is randomly chosen every \qty{10}{\second} - besides, when an obstacle is detected closer than \qty{40}{\centi\meter}, the direction is not changed to avoid turning back once the obstacle is almost avoided and staying longer than necessary close to obstacles. More details can be found in \cite{muller2024batdeck}.

\section{Velocity estimation algorithm}
We first give an overview of the developed velocity estimation algorithm and then provide the mathematical intuition for the used formula.
\subsection{Algorithm}
\begin{figure}
\centering
\includegraphics[width=0.4\columnwidth]{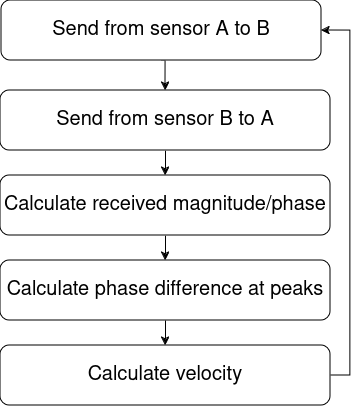}
\caption{Algorithmic flow of the ego-velocity estimation.}
\label{fig:flow_vel}
\end{figure}
The algorithm flow is shown in \Cref{fig:flow_vel}, and it first starts by sending sound from sensor A to sensor B and then the other way around, as visualized in \Cref{fig:maths_all}. Depending on the velocity of the drone, the traveled distance of the ultrasonic signal changes. The length difference in the paths can be translated to a phase difference in the received signals. By calculating the magnitude and phase out of the IQ data given by the sensor we can find the phase difference at the magnitude peak (which corresponds to the reflection from the ground).

\subsection{Mathematical intuition for velocity estimation}
\begin{figure}
    \centering
    \includegraphics[width=0.65\linewidth]{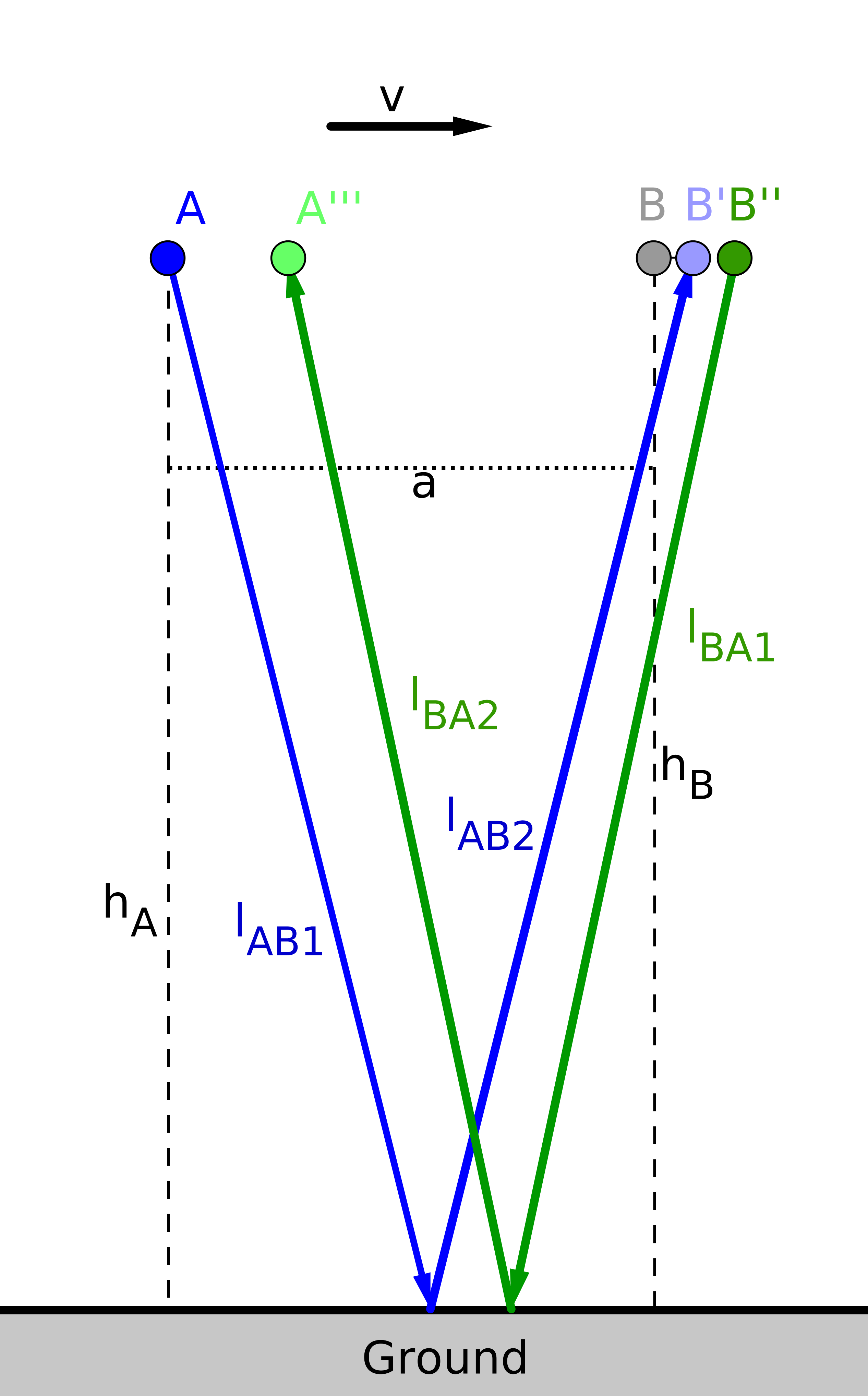}
    \caption{The drone features two sensors (A and B) $a$ apart. Sensor A sends a ultrasonic pulse at $t_A$, which sensor B receives at $t_{B'}$. Similarly, sensor B sends at $t_{B''}$, with sensor A receiving at $t_{A'''}$,. The drone flies with a constant velocity $v$ in positive x direction and the sensors are at heights $h_A$ and $h_B$.}
    \label{fig:maths_all}
\end{figure}

This section explains the mathematical background of the formula used to estimate a 1-d ego-velocity from two ultrasonic measurements reflected by the ground. It is based on a \gls{tof} difference, or signal travel length difference, measured between two sensors, depending on the speed of the drone.

To compensate for height differences between the sensors and the ground, which could be caused by the drone's pitch during flight, we send one ultrasonic pulse from sensor A and receive with sensor B and a second pulse from sensor B to sensor A, as shown in \Cref{fig:maths_all}. 
Furthermore, within this pulse sequence, we assume: 
\begin{align}
h_{A} & \approx h_{A'''} &  h_{B'} & \approx h_{B''} & t_{AB1} \approx t_{BA2} \approx t_1 
\end{align}
To calculate the difference in traveled length: $\Delta_{(m)}$, we can measure the phase difference of the signal arriving at sensor A and sensor B, which we denote as $\Delta_{(rad)}$, The difference in traveled length then calculates to: $\Delta_{(m)} = \Delta_{(rad)}\frac{\lambda}{2\pi}$.

\begin{figure}
    \centering
    \includegraphics[width=0.55\linewidth]{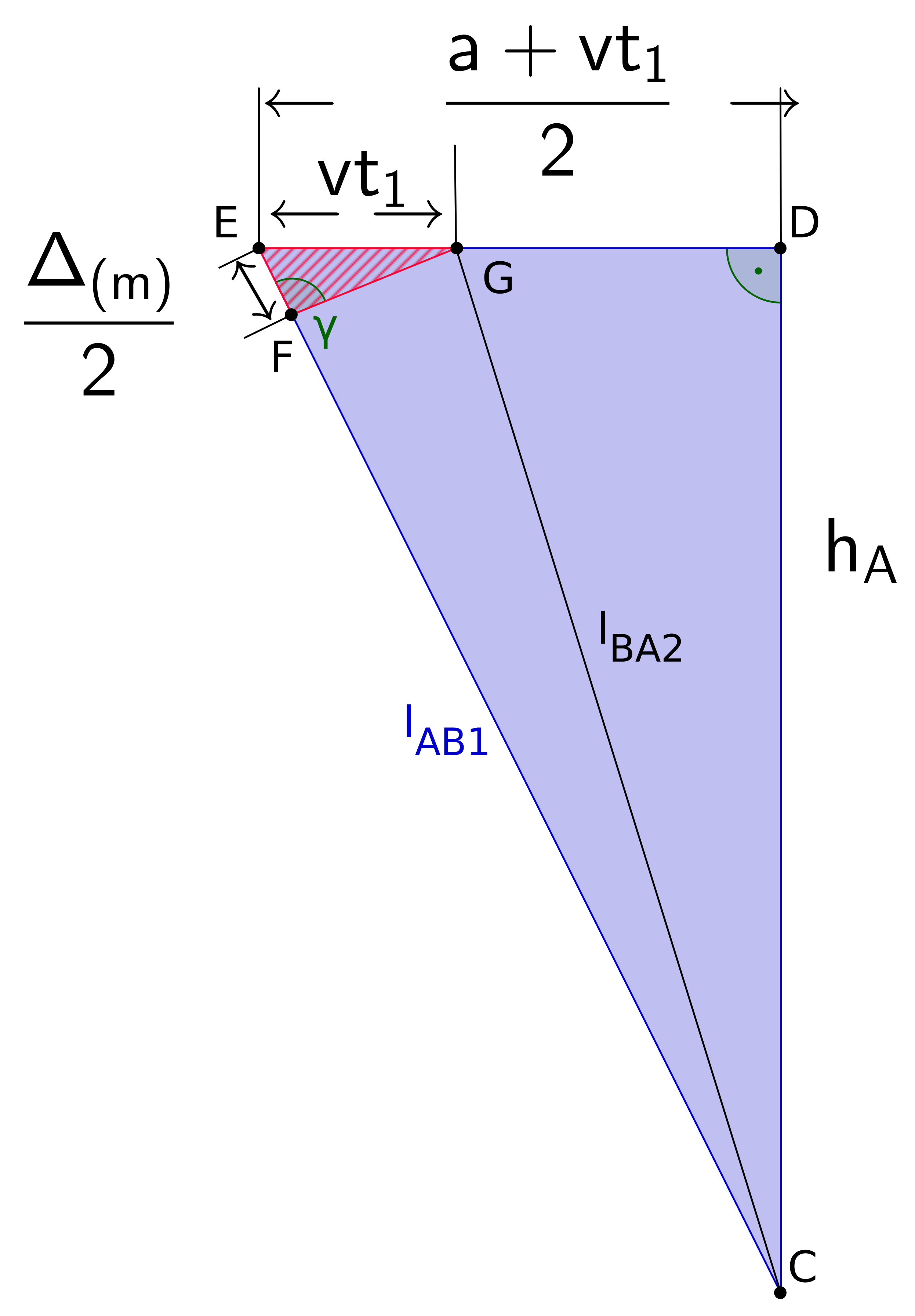}
    \caption{Here we show the intuition behind the formula while only looking at left (sensor A) side.}
    \label{fig:maths_all2}
\end{figure}

To estimate $v$ from  $\Delta_{(m)}$, we re-arrange \Cref{fig:maths_all} and draw $l_{AB1}$ and $l_{BA2}$ to meet at one point at the ground, denoted $C$. Looking only at the left part of the figure, we can visualize the traveled distance of the drone between sending and receiving as well as the difference between $l_{AB1}$ and $l_{BA2}$ (denoted as $\Delta_{(m)}/2$) in \Cref{fig:maths_all2}.

 The triangles $\overline{\rm CDE}$ and $\overline{\rm EFG}$ can be approximated as similar, as the angle $\gamma$ is close to $90\degree$ (as $vt << h_A$), and hence the ratio in \Cref{deqn_ex1} holds.
\begin{equation}
\label{deqn_ex1}
\frac{\overline{\rm EF}}{\overline{\rm EG}} = \frac{\overline{\rm DE}}{\overline{\rm CE}} \Leftrightarrow \frac{\Delta_{(m)}}{2vt_1} = \frac{a + vt_1}{2l_{AB1}}.
\end{equation}

\begin{align}
\label{deqn_ex2}
l_{AB1} &= \frac{c_0t_{1}}{2}
\end{align}
With the pulse travel distance in \Cref{deqn_ex2} and $c_0$ being the speed of sound, we can solve \Cref{deqn_ex1} which results in \Cref{deqn_v_complete}:
\begin{equation}
\label{deqn_v_complete}
v = \frac{-2a  \pm \sqrt{4a^{2} + 8t_1c_0\Delta_{(m)}}}{4t_1}.
\end{equation}
we choose the solution closer to $0 m/s$, which has to be the one with a positive square root (as $a$ is a positive distance and the result of a square root is $\geq 0$).
\\
Given this solution, we can approximate the velocity $v$ of the drone in \Cref{deqn_v_approx}.
\begin{align}
\label{deqn_v_approx}
a &>> vt_1  \implies &v &\approx \frac{\Delta_{(m)}c_0}{2a}
\end{align}
Notably, the height has no significant influence on the velocity calculation, as long as the approximation $a >> vt_1$ holds. This enables the use of the formula irrespective of the height at which the drone flies.
 The sensor distance $a$ defines the velocity measurement range, leading to a measurement range of $v \in (\SI{-4.5}{\mps},\SI{4.5}{\mps})$ for our setup. As the Crazyflie can maximally fly \SI{2}{\mps}, this range is more than sufficient; we can even filter out outliers that are faster than \SI{2}{\mps} without them being interpreted as false velocities because of phase wraps.

\section{Experimental Setup}
We use two separate physical setups to test \gls{oa} and the velocity estimation. All setups are based on the TKD ICU-x0201 ultrasonic sensors attached to our custom \textit{BatDeck} mounted on the \gls{cf} platform. \\
The two setups consist of:  
\begin{enumerate*}[label=(\roman*)]
\item For \gls{oa}: We mount one ICU-30201 sensor facing forward, as shown in the upper part of \Cref{fig:hw_overview}. 
\item For velocity estimation we mount two ICU-10201 sensors (A and B, $a = \SI{3.8}{\centi\meter}$ apart) under the drone (lower part of \Cref{fig:hw_overview}).
\end{enumerate*}

\begin{figure}
    \centering
    \includegraphics[width=0.9\linewidth]{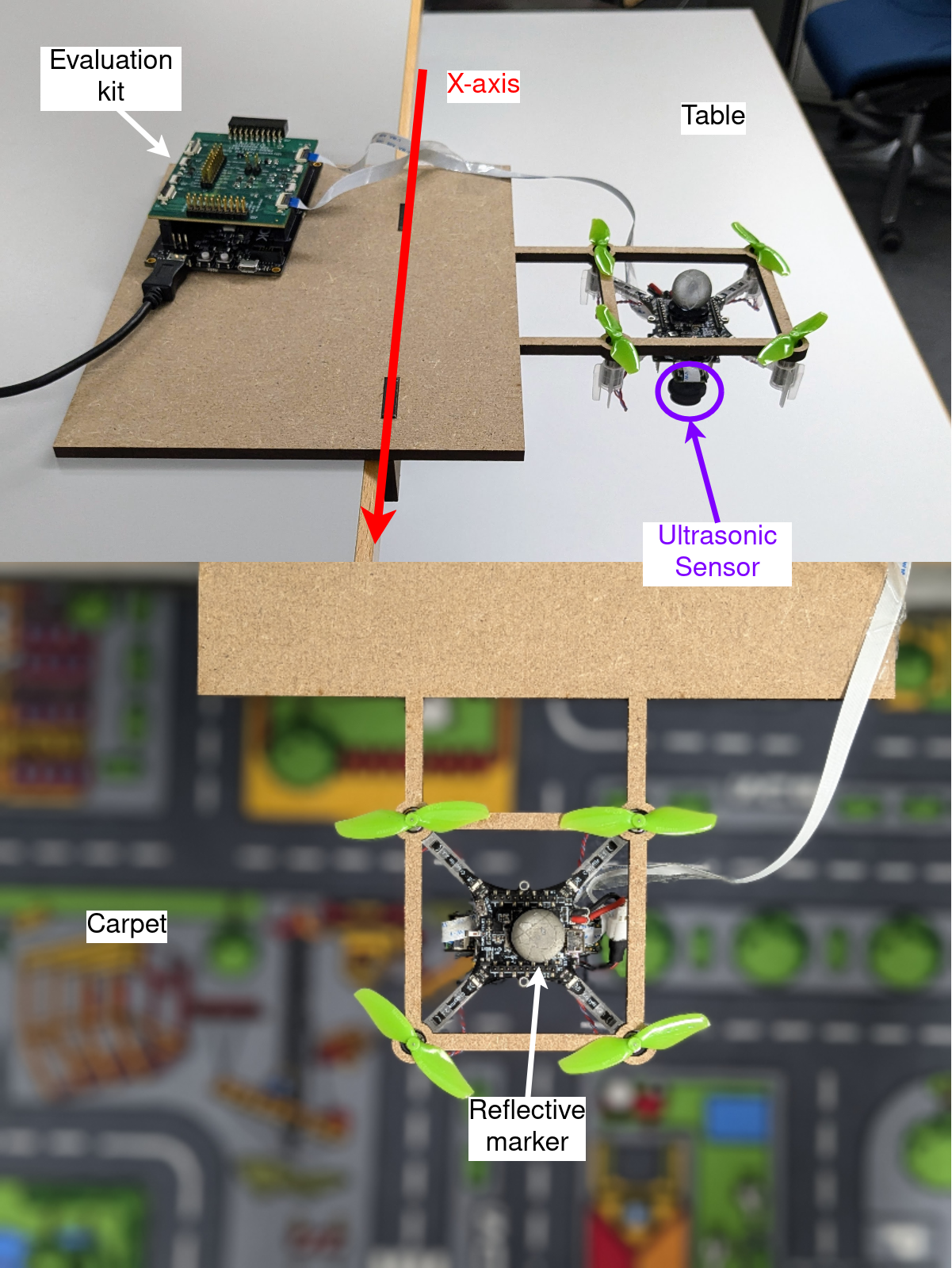}
    \caption{The measurement setup on a table features an evaluation kit for recording the ultrasonic data at full resolution, as well as a Crazyflie drone with a Flow-deck v2 and two downward-facing ultrasonic sensors. On top, a reflective marker is mounted, to acquire ground truth data with a motion capture system. We use two setups: over a smooth, featureless surface (top) and over a rough, feature-rich surface (bottom)}
    \label{fig:setup_table}
\end{figure}

For this velocity estimate proof-of-concept setup, we additionally built a stand for the drone that limits the movement to the X-axis and allows to turn on the motors (and lift the drone slightly) while the evaluation board for extensive dataset collection is still connected, as shown in \Cref{fig:setup_table}. Measurements were performed on a flat, optical feature-less table surface, as well as on an optical feature-rich carpet. The velocity ground truth is obtained by recording the drone's movements with a motion capture system.

\section{Obstacle Avoidance Results}

In this section, we evaluate the power consumption, latency, and computational load of the ICU-30201 sensor and the proposed \gls{oa} algorithm on the Cortex-M4 core of the \gls{cf}. We then describe the in-field performance of the \gls{oa} algorithm with the ICU-30201 sensor and compare it to a low-power laser-based \gls{tof} sensor (VL53L1) for the same task, including an analysis of failure modes for both sensors. 

\subsection{Power, Latency, and Computational Load}

To cover twice the maximum range of 340 samples at an $\gls{odr} = f_{op}/4$ in air (about \qty{4.6}{\meter}), the ultrasonic wave requires approximately \qty{27}{\milli\second}. The SPI data transfer for the complete \gls{iq} data (using \qty{12}{\mega\hertz} SPI) takes \qty{1.4}{\milli\second}. An additional \qty{3}{\milli\second} is needed for initial sensor communication before readout, including FreeRTOS scheduling delays with a variance of about ±\qty{1}{\milli\second}. Execution of the \gls{oa} algorithm requires under \qty{1}{\milli\second}. 

The next measurement can begin during the SPI transfer and \gls{oa} algorithm execution, leading to a total latency of approximately \qty{30}{\milli\second} per measurement or a rate of \qty{33}{\hertz}. Given the physical limits on transmit and receive times (which could only be reduced by shortening the signal duration and range), the only reducible delay for increasing the measurement rate is the initial \qty{3}{\milli\second} sensor communication time.

The \gls{oa} task adds only \qty{2.5}{\percent} to the STM32F405's computational load for the flight controller (\qty{41.5}{\percent}), leaving over \qty{57}{\percent} of processing capacity idle. The ICU-30201 sensor consumes less than \qty{1}{\milli\watt}, a negligible amount compared to the drone's total power usage of approximately \qty{11}{\watt}.

\subsection{In-field \gls{oa} Tests}
We performed 10 test flights in an office environment to validate our setup. The space included various obstacles, such as desks, chairs, cupboards, glass doors, and additional items like a cloth rack and a bicycle. The results, as summarized in \Cref{tab:oa_us}, show that the drone successfully completed its flight until the battery was depleted in \qty{50}{\percent} of the trials, with crashes occurring in the other flights. Most crashes resulted from low-altitude maneuvers beneath tables or collisions with soft obstacles, such as chairs. On average, each flight covered \qty{68}{\meter} with a flight time of \qty{4}{\minute} \qty{22}{\second}. Excluding battery changes, the average time and distance before a crash were \qty{8}{\minute} \qty{43}{\second} and \qty{136}{\meter}, respectively.

\begin{table}[t]
\caption{Results of 10 random exploration \gls{oa} flights in an office environment~\cite{muller2024batdeck}.}
\label{tab:oa_us}
\centering
\footnotesize
\begin{tabular}{c|c|c|c}
\toprule
{\textbf{Exp.}} & \textbf{Time [s]} & {\textbf{Crash}} & \textbf{Distance [m]} \\ 
\midrule
(1) & 430 & $\times$    & 107  \\ \hline
(2) & 294 & \checkmark  & 72  \\ \hline
(3) & 392 & $\times$    & 120  \\ \hline
(4) & 394 & $\times$    & 115  \\ \hline
(5) & 188 & \checkmark  & 55  \\ \hline
(6) & 287 & $\times$    & 63  \\ \hline
(7) & 35 & \checkmark   & 8.5  \\ \hline
(8) & 97 & \checkmark   & 27  \\ \hline
(9) & 383 & $\times$    & 90  \\ \hline
(10) & 115 & \checkmark & 27  \\ 
\midrule
\textbf{Average} & \textbf{262} & \textbf{50\%} & \textbf{68} \\ \bottomrule
\end{tabular}
\end{table}

\subsection{Comparison to VL53L1}
We conducted the real-world \gls{oa} test using the VL53L1 sensor in place of the ICU-30201, maintaining the same operating frequency (\qty{33}{\hertz}) and avoidance algorithm. Of the 10 flights, \qty{100}{\percent} resulted in crashes, with an average travel distance of \qty{4}{\meter} and a duration of \qty{9}{\second}. Notably, the VL53L1 sensor is disadvantaged by its difficulty in detecting highly reflective or absorptive surfaces, such as glass and black objects, and by its narrower \gls{fov} (\qty{27}{\degree} compared to \qty{55}{\degree} of the ICU-30201), which limits its ability to detect surrounding obstacles. Most crashes were due to undetected glass doors.

\begin{figure}.
\centering
\includegraphics[width=1\columnwidth]{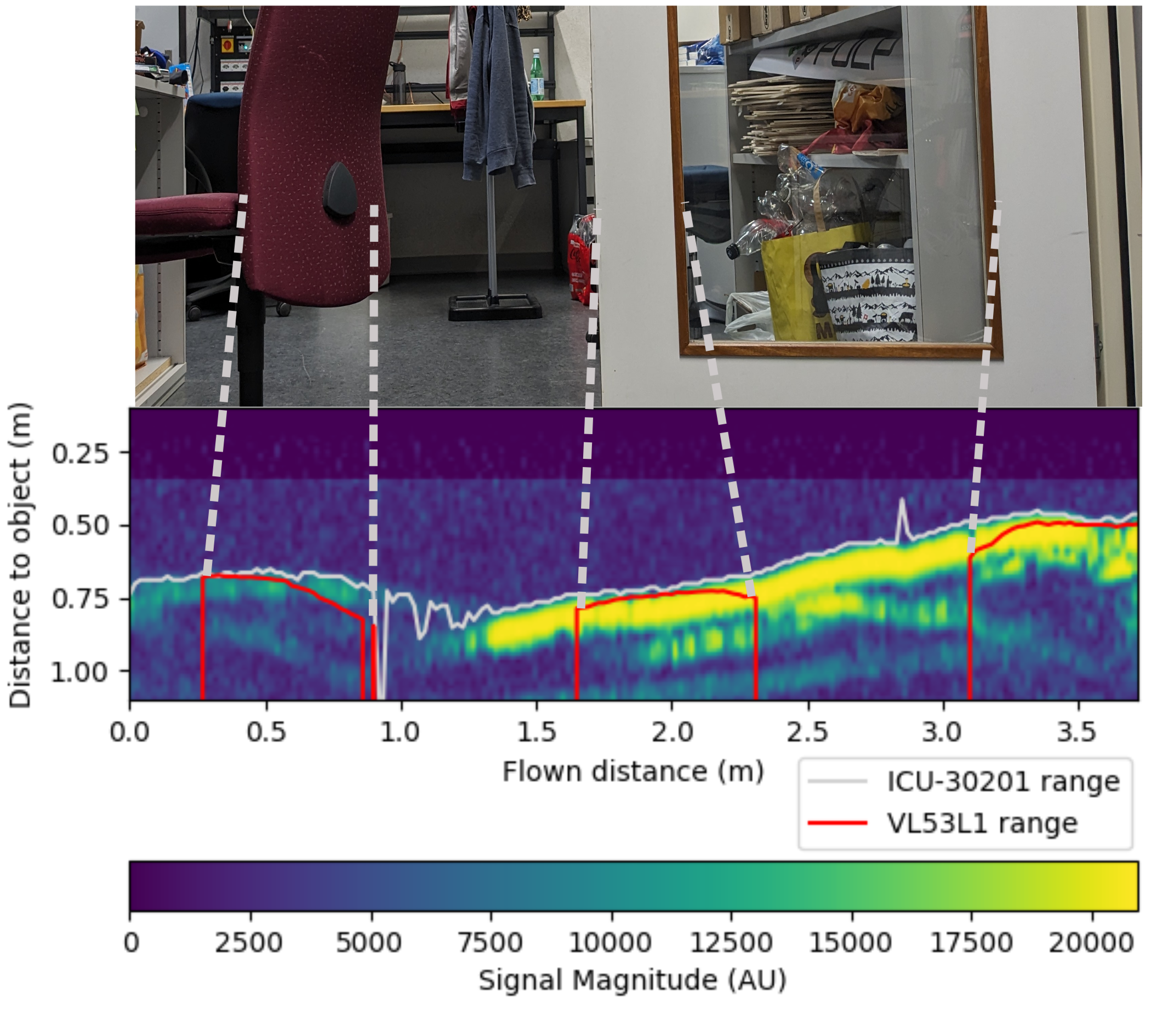}
\caption{ICU-30201 and VL53L1 comparison, both concurrently logged while flying at \qty{0.1}{\mps} from the left to the right while facing the obstacles shown in the top part. The correspondences between the VL53L1 and the image are highlighted with grey dashed lines~\cite{muller2024batdeck}.}
\label{fig:comp}
\end{figure}

To showcase the different perceptions of the two sensors and investigate common causes of failure, we flew along a row of different objects/materials, as displayed in \Cref{fig:comp}. Under the image of the scene, we show the ultrasonic spectrogram and the predicted distances to the closest object from both the ICU-30201 and the VL53L1 sensors. We see the effect of the narrower \gls{fov} of the VL53L1, enabling it to see narrower gaps but also provoking a more risky flight behavior. 
The VL53L1 fails to see the glass door and the first part of the chair, where it does not cover much of the \gls{fov}. The ICU-30201 successfully detects the soft chair, but it receives a much weaker reflection from it than from solid obstacles, especially when it is only at the edge of the \gls{fov}.    

In terms of power consumption, the VL53L1 consumes $\sim$\qty{50}{\milli\watt}, while the ICU-30201 consumes $<$\qty{1}{\milli\watt}. However, the \textit{BatDeck} with one sensor weighs \qty{3}{\gram}, while the Multi-ranger deck used for the Vl53L1 experiments reduced to 1 sensor weighs only \qty{2}{\gram}. Both designs can accommodate up to 4 sensors and, therefore, could be optimized for less weight for this \gls{oa} task.

\section{Velocity Estimation Results}
In this section, we first evaluate the power consumption of the ICU-10201 and the latency and computational load of the proposed ego-velocity estimation algorithm running onboard the \gls{cf} on the Cortex-M4 core. Following, we describe the ego-velocity evaluation with the ICU-10201 sensor and finally compare the results to the performance of a camera-based optical flow sensor (PMW3901) employed for the same task, also investigating causes for failure in both sensors. 
\subsection{Power, Latency and Computational Load}
The proposed sensor system consumes less than \SI{2}{\milli\watt}.
Measurement latency depends on the maximum height \(h\), calculated as $t_{latency} = \frac{4 \cdot h}{c_0}$
where \(c_0\) is the speed of sound. For a height of \SI{1}{\meter} above ground, latency is around \SI{12}{\milli\second}, with an additional \SI{3}{\milli\second} for communication, resulting in a maximum frequency of \SI{66}{\hertz}. The signal processing algorithm, which involves peak detection and computation using \Cref{deqn_v_complete}, requires negligible time, even on a Cortex-M4 core, ensuring efficient real-time performance.

The PMW3901 operates by calculating the optical flow of the floor below and depends on the \gls{tof} sensor VL53L1 for distance estimation to the ground and scaling of the optical flow to velocity. This sensor combination consumes approximately \SI{66}{\milli\watt} (\SI{18}{\milli\watt} for the PMW3901 and \SI{48}{\milli\watt} for the VL53L1) during operation at \SI{124}{FPS}. It returns the movement in 1/10 of a pixel leading to coarse resolution in higher heights.

\subsection{Ego-velocity Tests and Comparison to PMW3901}
\begin{figure}
    \centering
    \includegraphics[trim={0 0 6cm 0},clip, width=1.0\linewidth]{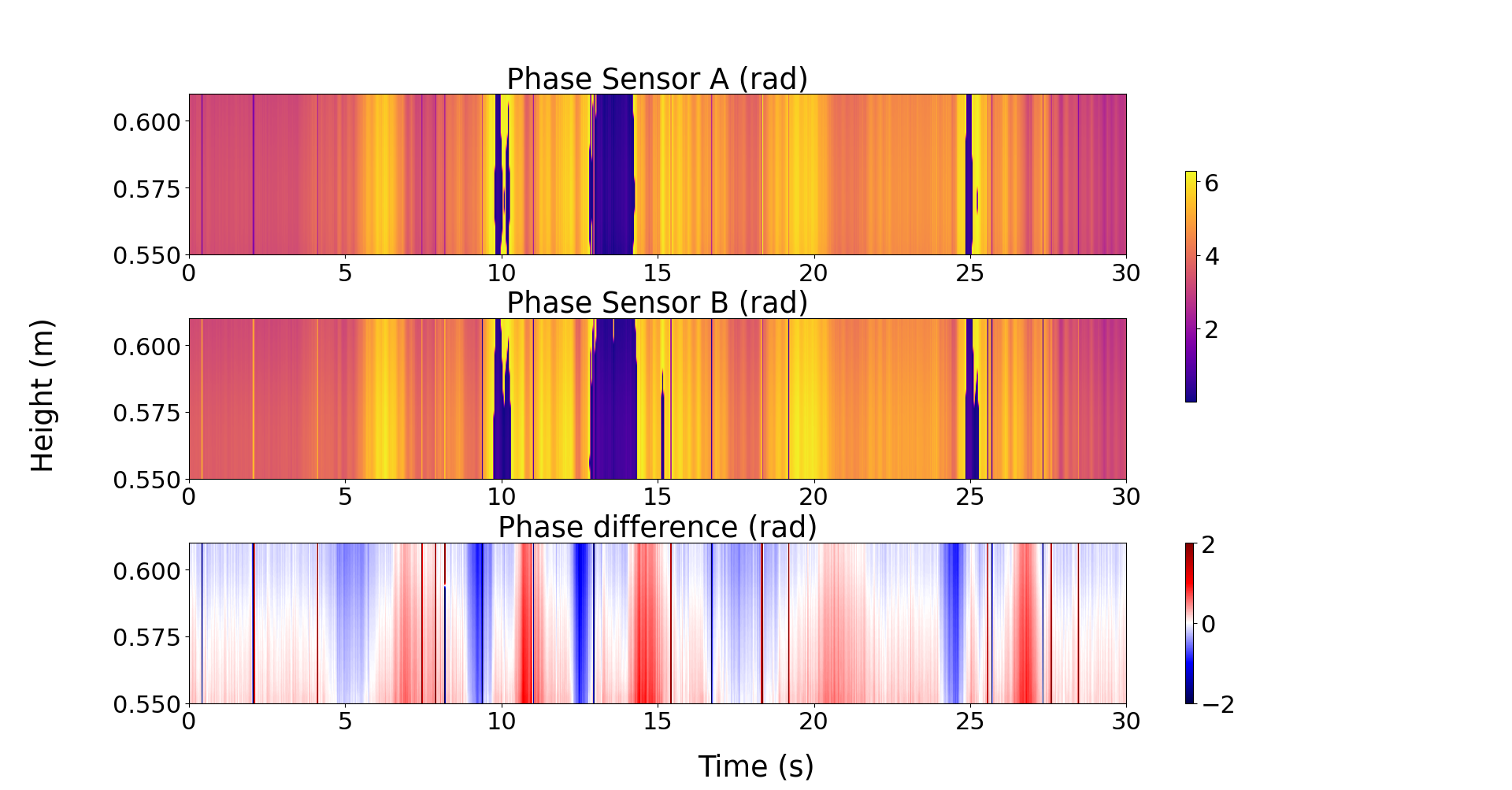}
    \caption{The phase difference (bottom) between sensor A (top) and sensor B (middle) corresponds to the X-velocity. These results were acquired on a flat, feature-less surface, in \SI{0.56}{\meter} height}
    \label{fig:results_phase_table}
\end{figure}
\begin{figure*}[b]
    \centering
    \includegraphics[width=0.8\textwidth]{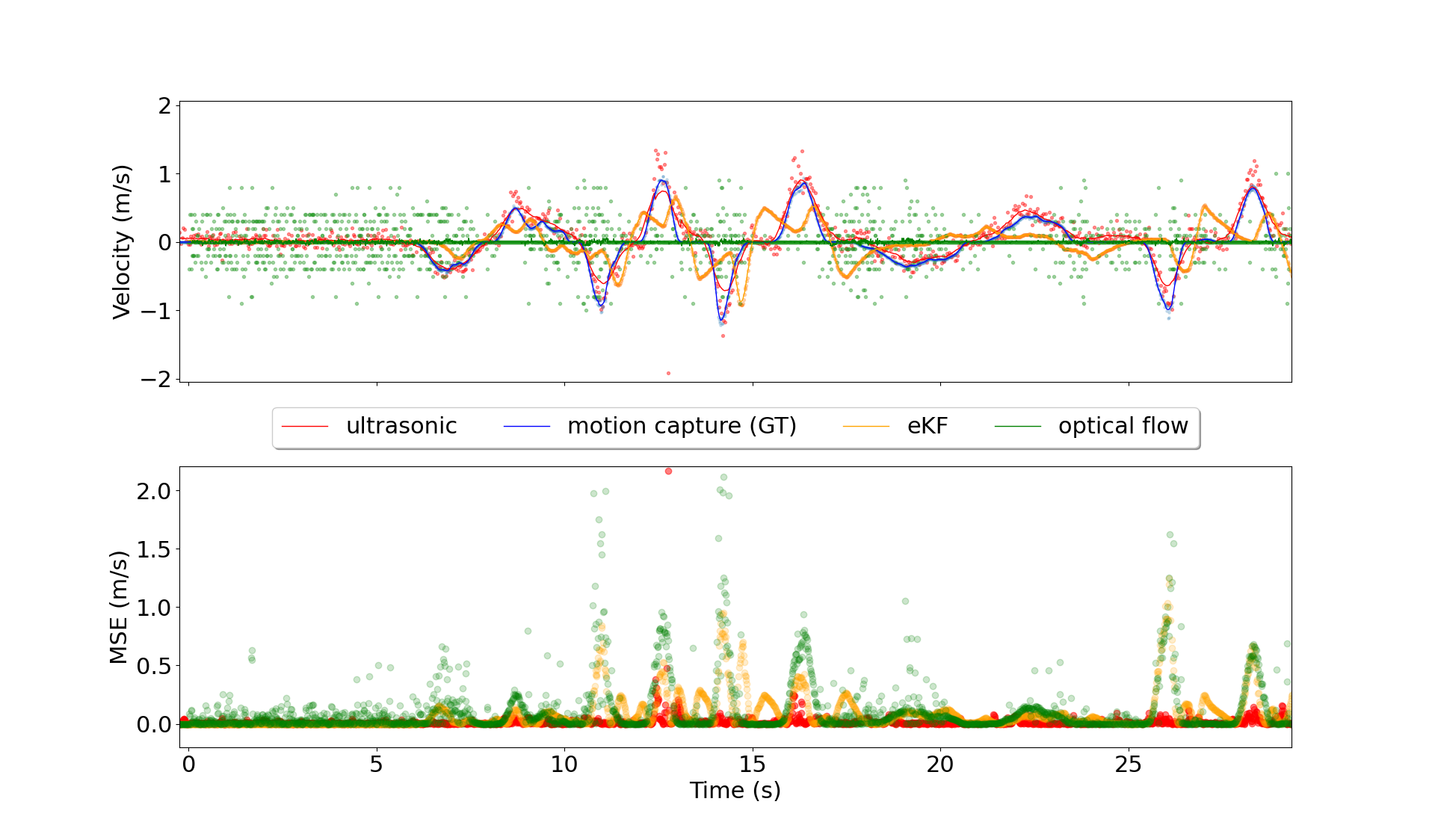}
    \caption{Top: The ground truth (GT) is shown in blue, our method in red, the raw optical flow measurements in green and the \gls{imu} fused with the optical flow measurements is shown in orange. Dots represent measurements, while lines show a moving average for better readability. These results were acquired on a flat surface, in \SI{0.56}{\meter} height. Bottom: The corresponding MSE of the measurements, outliers over \SI{2}{\meter\per\second} are not shown for readability.}
    \label{fig:res_table}
\end{figure*}

We acquired motion-capture ground truth data, raw optical flow estimation, fused (through an \gls{ekf}) optical flow by the \gls{cf} firmware, as well as \gls{imu} data and ultrasonic measurements at the same time.

In \Cref{fig:results_phase_table}, we show example phase measurements at sensor A, respectively B, and the resulting difference over multiple samples. This difference can then be used with \Cref{deqn_v_complete} to compute the velocity. The figure already gives an estimate of movement, with the phase difference being colored blue for negative speed and red for positive speed of the drone.

\begin{figure}
    \centering
    \includegraphics[trim={0 0 6cm 0},clip, width=1.0\linewidth]{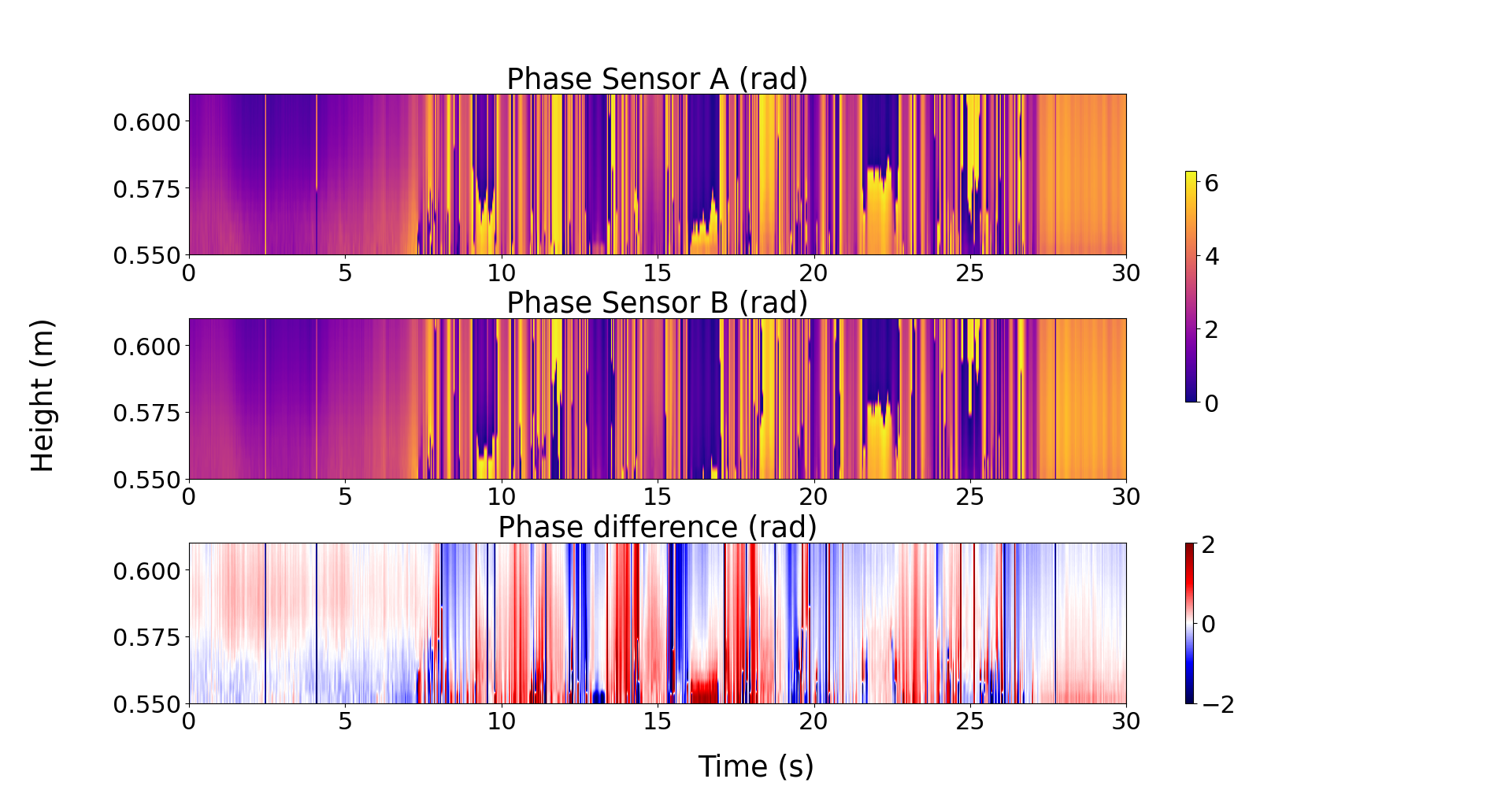}
    \caption[width=\linewidth]{The phase difference (bottom) between sensor A (top) and sensor B (middle) corresponds to the X-velocity. These results were acquired on a carpet, meaning a rough surface with optical features, in \SI{0.55}{\meter} height.}
    \label{fig:results_phase_carpet}
\end{figure}

We repeated the measurements over a carpet, which is challenging for the ultrasonic sensors as the surface is rough and the reflection therefore noisy, as shown in the phase plots in \Cref{fig:results_phase_carpet}. For the optical flow sensor, the feature-rich carpet is an ideal surface, as can be seen in the comparison with GT and ultrasonic data in \Cref{fig:res_carpet}. 

Based on these measurements, in \Cref{fig:res_table}, we show the resulting velocities, compared to the ground truth (motion capture system), the raw optical flow measurements and the \gls{ekf} (\gls{imu} and optical flow) results.
In \Cref{fig:res_table}, we show the measured/estimated velocity along the X-axis.

\begin{table}
    \centering
    \caption{\gls{mse} of the ultrasonic measurement, the optical flow measurement and the \gls{imu} fused with optical flow through an \gls{ekf}. }
    \label{tab:results_vel}
    \begin{tabular}{c|c|c}
    \toprule
        \textbf{Method} & \textbf{MSE Table (m/s)} & \textbf{MSE Carpet (m/s)}\\
    \midrule
        ultrasonic & \textbf{0.019} & 0.119\\ \hline
        optical flow & 0.112 & 0.009\\ \hline 
        eKF & 0.082 & \textbf{0.001}\\ \bottomrule
    \end{tabular}
\end{table}

We see that on a flat, feature-less surface (table), the proposed ultrasonic method outperforms the optical flow measurement, even if the latter is fused with \gls{imu} data. However, on a carpet the optical flow outperforms even the ultrasonic performance on a flat table.

\section{Conclusion}
We presented and evaluated the \textit{BatDeck}, featuring a novel, lightweight, and low-power ultrasonic sensor for nano-drones. We show the increased navigation robustness due to \gls{us} sensing in two use cases: \gls{oa} and ego-velocity estimation. 

\noindent%
\begin{minipage}{\textwidth}% to keep image and caption on one page
\makebox[\textwidth]{%        to center the image
  \includegraphics[width=0.8\textwidth]{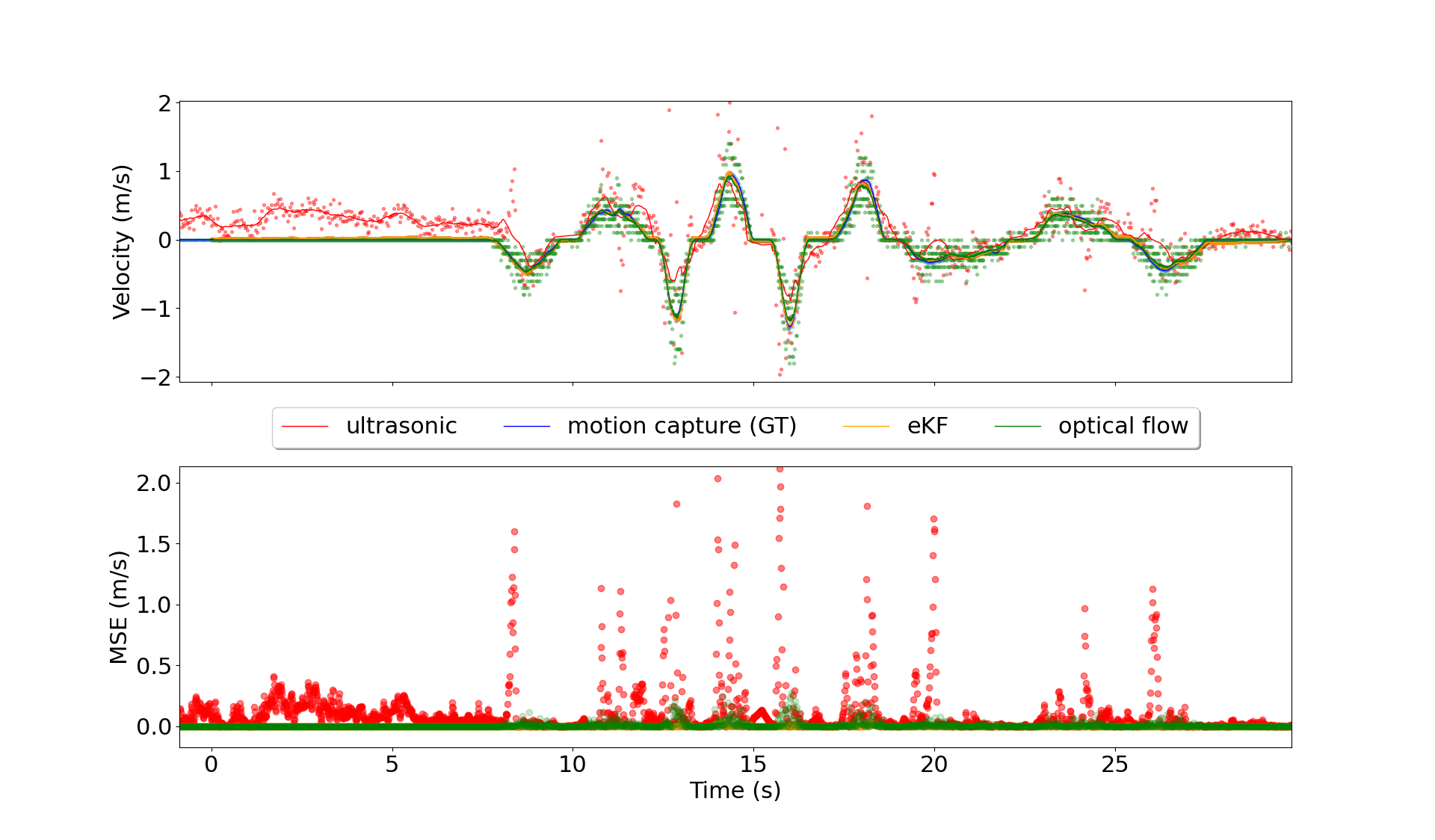}}
\captionof{figure}{Top: The ground truth (GT) is shown in blue, our method in red, the raw optical flow measurements in green and the \gls{imu} fused with the optical flow measurements is shown in orange. Dots represent measurements while lines show a moving average for better readability. These results were acquired on a carpet, meaning a rough surface with optical features, in \SI{0.55}{\meter} height. Bottom: The corresponding MSE of the measurements, outliers over \SI{2}{\meter\per\second} are not shown for readability.}\label{fig:res_carpet}

\end{minipage}

\newpage
We show the \gls{oa} capabilities by implementing a proof-of-concept algorithm that can run at \qty{33}{\hertz} and achieves an average flight distance of \qty{136}{\meter} and time of \qty{8}{\minute} \qty{43}{\second} until crash. Our ultrasonic \gls{oa} algorithm outperforms a classical laser-based \gls{tof} sensor, offering \qty{50}{\percent} higher mission success rate while tracking \qty{17}{\times} more distance. However, we also show the current limitations in detecting narrow gaps and soft obstacles.

We demonstrated that ego-velocity estimation is possible by implementing a novel measurement scheme that allows us to estimate the velocity from the phase shift between two received signals. We validate this approach in a 1-dimensional test setup and compare it with a conventional optical flow-based sensor, achieving a \qty{5}{\times} lower \gls{mse} on a flat, featureless surface.

Currently, the approach is limited by strong airflow below the drone during flight, which we aim to overcome in future work -- possible solutions are a more advanced pulse-sending scheme, which interleaves data sent from both sensors and hence passes through the same noise, a mechanical solution to avoid strong airflow in the measurement area or a machine learning-based solution which learns the effect of the motors. 

\newpage
Additionally, we aim to pursue adding multiple ultrasonic sensors, which will allow a full spatial perception of obstacles and, with this, more advanced \gls{oa} algorithms. Also, using phase information and modulating the emitted ultrasonic wave hold potential for improved spatial perception. For ego-velocity estimation, adding multiple ultrasonic sensors also bears the potential to generalize the approach to two dimensions, and by installing a whole array, to reduce noise and increase the measurement range.

Lastly, we plan to explore sensor fusion techniques to integrate information from ultrasonic, laser-ranged \gls{tof} sensors and optical flow sensors to increase the robustness of autonomous nano-drones.

\section*{Acknowledgments}
We thank Christian Vogt for his advice and help.
We also thank TDK for their support.

\bibliographystyle{IEEEtran}
\bibliography{bib,bib_new}

\begin{thebibliography}{10}
\providecommand{\url}[1]{#1}
\csname url@samestyle\endcsname
\providecommand{\newblock}{\relax}
\providecommand{\bibinfo}[2]{#2}
\providecommand{\BIBentrySTDinterwordspacing}{\spaceskip=0pt\relax}
\providecommand{\BIBentryALTinterwordstretchfactor}{4}
\providecommand{\BIBentryALTinterwordspacing}{\spaceskip=\fontdimen2\font plus
\BIBentryALTinterwordstretchfactor\fontdimen3\font minus \fontdimen4\font\relax}
\providecommand{\BIBforeignlanguage}[2]{{%
\expandafter\ifx\csname l@#1\endcsname\relax
\typeout{** WARNING: IEEEtran.bst: No hyphenation pattern has been}%
\typeout{** loaded for the language `#1'. Using the pattern for}%
\typeout{** the default language instead.}%
\else
\language=\csname l@#1\endcsname
\fi
#2}}
\providecommand{\BIBdecl}{\relax}
\BIBdecl

\bibitem{fortune_business_insights_2024}
\BIBentryALTinterwordspacing
F.~B. Insights, ``Commercial drone market report,'' 2024, accessed: 2024-11-11. [Online]. Available: \url{https://www.fortunebusinessinsights.com/commercial-drone-market-102171}
\BIBentrySTDinterwordspacing

\bibitem{HASSANALIAN201799}
\BIBentryALTinterwordspacing
M.~Hassanalian and A.~Abdelkefi, ``Classifications, applications, and design challenges of drones: A review,'' \emph{Progress in Aerospace Sciences}, vol.~91, pp. 99--131, 2017. [Online]. Available: \url{https://www.sciencedirect.com/science/article/pii/S0376042116301348}
\BIBentrySTDinterwordspacing

\bibitem{macario2022comprehensive}
A.~Macario~Barros, M.~Michel, Y.~Moline, G.~Corre, and F.~Carrel, ``A comprehensive survey of visual slam algorithms,'' \emph{Robotics}, vol.~11, no.~1, p.~24, 2022.

\bibitem{he2020review}
M.~He, C.~Zhu, Q.~Huang, B.~Ren, and J.~Liu, ``A review of monocular visual odometry,'' \emph{The Visual Computer}, vol.~36, no.~5, pp. 1053--1065, 2020.

\bibitem{tim2022egomotion}
H.~Yin, P.~X. Liu, and M.~Zheng, ``Ego-motion estimation with stereo cameras using efficient 3d–2d edge correspondences,'' \emph{IEEE Transactions on Instrumentation and Measurement}, vol.~71, pp. 1--11, 2022.

\bibitem{tim2021monocular}
X.~Ban, H.~Wang, T.~Chen, Y.~Wang, and Y.~Xiao, ``Monocular visual odometry based on depth and optical flow using deep learning,'' \emph{IEEE Transactions on Instrumentation and Measurement}, vol.~70, pp. 1--19, 2021.

\bibitem{oho2022ultrasonic}
S.~Oho and M.~Ohkuma, ``Ultrasonic differential odometry for vehicle localization,'' in \emph{2022 61st Annual Conference of the Society of Instrument and Control Engineers (SICE)}, 2022, pp. 809--814.

\bibitem{zhmud2018application}
V.~Zhmud, N.~Kondratiev, K.~Kuznetsov, V.~Trubin, and L.~Dimitrov, ``Application of ultrasonic sensor for measuring distances in robotics,'' in \emph{Journal of Physics: Conference Series}, vol. 1015, no.~3.\hskip 1em plus 0.5em minus 0.4em\relax IOP Publishing, 2018, p. 032189.

\bibitem{rshen2019new}
M.~Shen, Y.~Wang, Y.~Jiang, H.~Ji, B.~Wang, and Z.~Huang, ``A new positioning method based on multiple ultrasonic sensors for autonomous mobile robot,'' \emph{Sensors}, vol.~20, no.~1, p.~17, 2019.

\bibitem{tim2011indoor}
J.~Eckert, R.~German, and F.~Dressler, ``An indoor localization framework for four-rotor flying robots using low-power sensor nodes,'' \emph{IEEE Transactions on Instrumentation and Measurement}, vol.~60, no.~2, pp. 336--344, 2011.

\bibitem{przybyla2023mass}
R.~J. Przybyla, S.~E. Shelton, C.~Lee, B.~E. Eovino, Q.~Chau, M.~H. Kline, O.~I. Izyumin, and D.~A. Horsley, ``Mass produced micromachined ultrasonic time-of-flight sensors operating in different frequency bands,'' in \emph{2023 IEEE 36th International Conference on Micro Electro Mechanical Systems (MEMS)}.\hskip 1em plus 0.5em minus 0.4em\relax IEEE, 2023, pp. 961--964.

\bibitem{tim2022multigoal}
W.~Xiao, L.~Yuan, L.~He, T.~Ran, J.~Zhang, and J.~Cui, ``Multigoal visual navigation with collision avoidance via deep reinforcement learning,'' \emph{IEEE Transactions on Instrumentation and Measurement}, vol.~71, pp. 1--9, 2022.

\bibitem{10423569}
K.~Kalenberg, H.~Müller, T.~Polonelli, A.~Schiaffino, V.~Niculescu, C.~Cioflan, M.~Magno, and L.~Benini, ``Stargate: Multimodal sensor fusion for autonomous navigation on miniaturized uavs,'' \emph{IEEE Internet of Things Journal}, pp. 1--1, 2024.

\bibitem{mueller2023robust}
H.~Müller, V.~Niculescu, T.~Polonelli, M.~Magno, and L.~Benini, ``Robust and efficient depth-based obstacle avoidance for autonomous miniaturized uavs,'' \emph{IEEE Transactions on Robotics}, vol.~39, no.~6, pp. 4935--4951, 2023.

\bibitem{lamberti2022tiny}
L.~Lamberti, V.~Niculescu, M.~Barciś, L.~Bellone, E.~Natalizio, L.~Benini, and D.~Palossi, ``Tiny-pulp-dronets: Squeezing neural networks for faster and lighter inference on multi-tasking autonomous nano-drones,'' in \emph{2022 IEEE 4th International Conference on Artificial Intelligence Circuits and Systems (AICAS)}, 2022, pp. 287--290.

\bibitem{laurijssen2019flexible}
D.~Laurijssen, R.~Kerstens, G.~Schouten, W.~Daems, and J.~Steckel, ``A flexible low-cost biologically inspired sonar sensor platform for robotic applications,'' in \emph{2019 International Conference on Robotics and Automation (ICRA)}, 2019, pp. 9617--9623.

\bibitem{forouher2016sensor}
D.~Forouher, M.~G. Besselmann, and E.~Maehle, ``Sensor fusion of depth camera and ultrasound data for obstacle detection and robot navigation,'' in \emph{2016 14th International Conference on Control, Automation, Robotics and Vision (ICARCV)}, 2016, pp. 1--6.

\bibitem{muller2024batdeck}
H.~Müller, V.~Kartsch, M.~Magno, and L.~Benini, ``Batdeck: Advancing nano-drone navigation with low-power ultrasound-based obstacle avoidance,'' in \emph{2024 IEEE Sensors Applications Symposium (SAS)}, 2024, pp. 1--6.

\bibitem{shan2020probabilistic}
M.~Shan, J.~S. Berrio, S.~Worrall, and E.~Nebot, ``Probabilistic egocentric motion correction of lidar point cloud and projection to camera images for moving platforms,'' in \emph{2020 IEEE 23rd International Conference on Intelligent Transportation Systems (ITSC)}.\hskip 1em plus 0.5em minus 0.4em\relax IEEE, 2020, pp. 1--8.

\bibitem{MUMUNI2022100416}
\BIBentryALTinterwordspacing
F.~Mumuni, A.~Mumuni, and C.~K. Amuzuvi, ``Deep learning of monocular depth, optical flow and ego-motion with geometric guidance for uav navigation in dynamic environments,'' \emph{Machine Learning with Applications}, vol.~10, p. 100416, 2022. [Online]. Available: \url{https://www.sciencedirect.com/science/article/pii/S2666827022000913}
\BIBentrySTDinterwordspacing

\bibitem{park2021}
Y.~S. Park, Y.-S. Shin, J.~Kim, and A.~Kim, ``3d ego-motion estimation using low-cost mmwave radars via radar velocity factor for pose-graph slam,'' \emph{IEEE Robotics and Automation Letters}, vol.~6, no.~4, pp. 7691--7698, 2021.

\bibitem{palossi201964}
D.~Palossi, A.~Loquercio, F.~Conti, E.~Flamand, D.~Scaramuzza, and L.~Benini, ``A 64-mw dnn-based visual navigation engine for autonomous nano-drones,'' \emph{IEEE Internet of Things Journal}, vol.~6, no.~5, pp. 8357--8371, 2019.

\bibitem{zhang2024endtoend}
\BIBentryALTinterwordspacing
N.~Zhang, F.~Nex, G.~Vosselman, and N.~Kerle, ``End-to-end nano-drone obstacle avoidance for indoor exploration,'' \emph{Drones}, vol.~8, no.~2, 2024. [Online]. Available: \url{https://www.mdpi.com/2504-446X/8/2/33}
\BIBentrySTDinterwordspacing

\bibitem{barron1994performance}
J.~L. Barron, D.~J. Fleet, and S.~S. Beauchemin, ``Performance of optical flow techniques,'' \emph{International journal of computer vision}, vol.~12, pp. 43--77, 1994.

\bibitem{duembgen2023blind}
F.~Dümbgen, A.~Hoffet, M.~Kolundžija, A.~Scholefield, and M.~Vetterli, ``Blind as a bat: Audible echolocation on small robots,'' \emph{IEEE Robotics and Automation Letters}, vol.~8, no.~3, pp. 1271--1278, 2023.

\end{thebibliography}

\newpage

\vfill

\end{document}